\documentclass[review]{elsarticle}
\usepackage{multicol}
\usepackage{multirow}
\usepackage{array}
\usepackage{ulem}
\usepackage[dvipsnames]{xcolor}
\usepackage{mdwmath}
\usepackage{mdwtab}
\usepackage{eqparbox}
\usepackage{url}
\usepackage{algorithmic}
\usepackage{textcomp}
\PassOptionsToPackage{unicode}{hyperref}
\usepackage{caption}
\usepackage{subcaption}
\usepackage{natbib}
\usepackage{graphicx}
\usepackage{colortbl}
\usepackage{url}
\usepackage{float}
\usepackage{booktabs}
\usepackage{color}
\usepackage{mathtools}
\usepackage{float}
\usepackage{mathrsfs}
\usepackage{comment}
\usepackage{amsmath}
\usepackage{amssymb}
\usepackage{listings}
\journal{Journal of \LaTeX\ Templates}

\begin{document}

\begin{frontmatter}

\title{Intrusion Detection using Spatial-Temporal features based on Riemannian Manifold}


\author[mymainaddress]{Amardeep Singh \corref{mycorrespondingauthor}}
\ead{a.singh1@massey.ac.nz}
\author[mymainaddress]{Julian Jang-Jaccard}
\cortext[mycorrespondingauthor]{Corresponding author}
\address[mymainaddress]{Cybersecurity Lab, Massey University, Auckland, NEW ZEALAND}

\begin{abstract}
Network traffic data is combination of different data bytes packet under different network protocols. These traffic packets have complex time-varying non-linear relationships. Existing state of the art method rise up to this challenge by fusing features into multiple subset based on correlations and using hybrid classification techniques that extract spatial and temporal characteristics. This offen requires high computational cost and manual support that limit them for real-time processing of network traffic. To address this, we propose a new novel feature extraction method based on covariance matrices that extracts spatio-temporal characteristics of network traffic data for detecting malicious network traffic behaviour. The covariance matrices in our proposed method not just naturally encodes the mutual relationships between different network traffic values but also have well defined geometry that falls in Riemannian manifold. Riemannian manifold is embedded with distance metrices that facilitates extracting descriminative features for detecting malicious network traffic. We evaluated our model on an NSL-KDD and UNSW-NB15 datasets and shown our proposed method significantly outperforms the conventional method and other existing studies on the dataset.
\end{abstract}


\end{frontmatter}

\section{Introduction}
\label{sec:introduction}
\begin{figure}[h]
   \centering
    \includegraphics[scale=0.55]{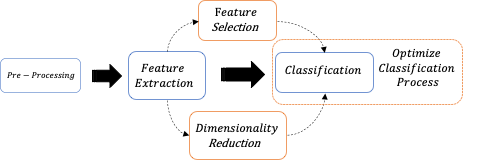}    
    \caption{Framework for ML based IDS}
    \label{fig:frameworkML}
\end{figure}

Intrusion detection systems (IDS) can distinguish between normal and abnormal network activity by monitoring data packets. This makes IDS a necessary part of computer network security infrastructure for almost all organizations.  
The current state of machine learning (ML) based on IDS consists of the preprocessing, feature extraction, and classification steps, as shown in Figure \ref{fig:frameworkML}. The ML pipeline for IDS suffers from complexity issues (large amount of high dimensional complex data) and is insufficient for learning complex nonlinear relationships that change over time between large datasets \cite{IERACITANO202051}. As Hogo \cite{6987012} stated in his work, the existing IDS mostly focus on the last snapshot, while there are temporal and spatial dependencies in the traffic data \cite{article4}.  Like, source IP and destination IP define the subject and object of the behavior in the data streams, and the duration describes how long the behavior lasted \cite{Rieman1}. Similarly, the packet volume and packet size indicate the traffic flow, and their size varies between different protocols \cite{Rieman1}. Therefore, this information should be analyzed in the context of the communication protocol  to explain the impact of this behavior on the communication capability. This spatio-temporal relationship between the multivariate data helps in detecting various attack  (such as DDos attacks, etc.) characteristics  that look rather benign under different network protocols \cite{Rieman1}.
Existing state of the art methods address this issue by proposing feature fusion and association to improve ability to identify malicious network behaviours \cite{li2018data}. For example, Li et. al \cite{LI2020107450} proposed IDS that divide features into four subset based on correlation among features and used multi-convolutional neural network to detect intrusion attacks. Similarly, Wei et al. \cite{8171733} proposed fusion of convolutional neural network (CNN) to learn spatial features and long short term memory (LSTM) to learn the temporal features among multiple network packets. As traffic data is combination of different data bytes packet under different network protocols. These network traffics show similar characteristics of natural language \cite{article5}. It is equivalent to finding sentiment analysis based on text whether it is positive or negative comments  \cite{article5}.  In past many feature fusion methods based on  end-to-end learning by combining recurrent neural network and CNN architecture \cite{article302} are used to extract spatial-temporal characteristic but these solutions depend upon finding right architecture and parameters. Sometime optimization techniques are required to find optimal parameters \cite{article5} which limits the performance for real time processing.\\
To achieve real-time recognition, the IDS pipeline should be simple enough. This can be best achieved if the feature extraction method able to find most descriminative information between normal and abnormal network traffic. To address this, we proposed an algorithm where feature extraction is based on covariance matrices of samples. Covariance matrices have proven to be powerful features to extract spatio-temporal characteristics  in a variety of applications, such as time series classification \cite{ergezer2018time}, network anomaly detection \cite{5485505}, and brain-computer interface \cite{1Alexandre}.
The sample covariance matrices (SCMs) consider mutual relationships between multivariate distributions \cite{9231898}. In this work, we not only exploit the mutual relationships between variables through SCMs, but also the geometry of covariance matrices. In general, covariance matrices form a Riemannian hyper-cone embedded with distance metric that facilitates discriminate information extraction in the manifold \cite{1Alexandre,9231898}. The main contributions of this paper can be summarised as follows.

\begin{itemize}
    \item We propose a new feature extraction method based on covariance matrices for learning spatial-temporal characteristics in network traffic data. 
    
    \item The covariance matrices we use in our feature extraction  falls in Riemannian manifold. Thus geometrical properties can be better utilized by using manifold based learning algorithm to improve classification accuracy in  detecting malicious network traffic.
    
    \item Our work is further extended to support Euclidean space classification algorithms (like almost all ML and DL methods) by mapping covariance matrices back to euclidean tangent space to Riemannian manifold.
    
    \item Our experimental results show that manifold based learning algorithm in the feature extraction strategy is highly effective in increasing detection accuracy. 
    
\end{itemize}
The rest of the paper is organized as follows. Section $II$ presents related works for intrusion detection. Section $III$ introduces our proposed IDS approach. Section $IV$, describes experimental setup and evaluation metrics. In section $V$, the experimental results are discussed and compared to existing studies. Lastly, section $VI$ draws the conclusions of our proposed method.
\section{Related Work}
The ML based IDS in the literature address issues like data complexity and non-linear relationships that change over time between large datasets by selecting most suitable features or parameters for classification algorithm though some  optimization technique. IDS based on this strategy in literature can be classified into: (a) supervised IDS based on shallow learning techniques, (b) unsupervised IDS and (c) Deep learning based IDS.
\subsection{Supervised Intrusion detection}
Supervised learning based IDS used labelled data to detect intrusion detect but require suitable parameter for dealing with complex traffic data. For example, Li et al  \cite{article339} proposed a model based on the Gradient Boosted Decision Tree (GBDT) algorithm. They used the particle swarm optimization (PSO) algorithm to optimise the parameters of the GBDT-based model for intrusion detection. Similarly, Papamartzivanos et al  \cite{article333} proposed an interpretable decision tree model for intrusion detection. They used a genetic algorithm to avoid biassing the decision tree towards the most abundant instances of the intruder class in the dataset. To find optimal parameters for ML models, Aljawarnehet al \cite{article330}  proposed a hybrid model combining different machine learning classifiers such as NaiveBayes, AdaBoostM1, Meta Pagging, J48, DecisionStump, REPTree, and RandomTree. On the same line, Dhaliwal et al \cite{article325} used ensemble learning method35 (generates a predictive model from an ensemble of weak ML models) based on gradient boosting framework (XGBoost algorithm) to detect intrusion attacks. Alabdallah and Awad \cite{article338} also proposed a stratified sampling and weighted Support Vector Machine (W-SVM) method that leads to improvement in low frequency attack detection in intrusion detection model. Yu et al. \cite{article328} presented an online adaptive model to meet the requirements of a real-time intrusion detection mechanism based on an extreme learning machine (ELM). The model is regularized using Tikhonov regularization to avoid overfitting. The adaptive mechanism is proposed to update the model weights and regularization parameter based on the received and available data. The results show faster training speed, higher accuracy and lower false positive/negative rate.\\
In addition to optimizing the classification process in the IDS pipeline, researchers also find an optimal subset of features to improve IDS performance. Such as Hajisalem and Babaie \cite{article327} proposed a method in which the training data is divided into subsets based on fuzzy C-means clustering, and then irrelevant features are removed based on correlation creteria. Finally, decision rules are generated for each subset using CART technique to classify normal and abnormal instances. Lastly, a hybrid of Artificial Bee Colony (ABC) and Artificial Fish Swarm (AFS) algorithms is used to generate final rules for intrusion detection.
In the same vein, Ghazy et al. \cite{article326} used wrapper-based feature selection to find the most relevant features for a random forest ensemble classifier for intrusion detection.
Latah and Toker \cite{article313} used the PCA approach for feature selection and dimensionality reduction along with bagging and boosting approaches for intrusion detection and classification.
Donkal and Verma \cite{article311} used non-dominated Sorting Genetic Algorithm to select the most promising features to be subjected to a combination of different machine learning models based on majority voting to detect and classify intrusion attacks.
In their work, Wang et al. \cite{article307}  eliminate redundant and irrelevant features based on Pearson correlation technique and then create a support vector machine model for intrusion detection from the reduced feature set.
Similarly, Gowsic et al. \cite{article300} wrapper-based feature selection to find the most relevant features for a Adaboost Decision tree mode for intrusion detection.
Iwendi et al.\cite{article12} proposed IDS which combines a genetic algorithm for feature optimization (finding the best feature set) with ensemble learning based method Random Forest (RF) to achieve high detection rate and low false alarm rate.
Whether it is selecting most suitable features or parameters for classification algorithm requires optimization which comes at the cost of time. This process might need to repeat due to time-varying relationships in data (e.g. concept drift), thus make it not suitable for real-time intrusion detection.
\subsection{Unsupervised Intrusion detection}
One of the main drawback of supervised methods is that they require labeled training data, which is very tedious task. Therefore, unsupervised methods have recently gained attention in the research community. For example, Auskalnis et al \cite{article343} use Local Outlier Factor during data preprocessing to exclude normal packets that overlap with the density position of anomalous packets. Later, a cleaned (reduced) set of normal packets is used to train another local outlier model to detect anomalous packets. Similarly, Rathore et al \cite{article314} proposed unsupervised IDS that is based on semi-supervised fuzzy C-Mean clustering with single hidden layer feedforward neural networks (also known as Extreme Learning Machine) to detect intrusions in real-time. Aliakbarisani et al \cite{article308} proposed method that learns a transformation matrix based on the Laplacian eigenmap technique to map the features of samples into a new feature space, where samples can be clustered into different classes using data-driven distance metrics.\\
With the advent of deep learning methods and cheap hardware (graphical processing units), unsupervised deep learning methods such as autoencoders and self-organizing maps are increasingly being used. Karami \cite{article315} proposed an IDS model based on a self-organizing map that not only removes benign outliers but also improves the detection of anomalous patterns. Similarly, Song et al \cite{article7} proposed an auto-encoder model (trained on normal samples) based on the principle that the reconstruction loss of normal traffic samples is lower than that of abnormal (attack) samples, so that a threshold can be set for detecting future attacks. In addition, this work evaluates various hyperparameters, model architectures, and latent size settings in terms of attack detection performance. Various researcher proposed methods that combine  unsupervised and supervised  methods to get best of both learning techniques. Like, Shone et al \cite{article331}  proposed unsupervised feature learning using a non-symmetric stacked deep autoencoder. Moreover, these features are used for intrusion detection using a random forest model. Similarly, Hawawreh et al \cite{article321} combined autoencoder with deep feed-forward neural network for intrusion detection.
\subsection{Deep learning based Intrusion detection}
Deep learning models (DL) offers end-to-end solutions that do not require manual feature extraction or selection and able to learn relationship between data values. Deep learning models contain various hyper parameters (learning rate, batch size, number and size of hidden layers, etc.) that affect the performance of the supervised and unsupervised DL model. Sometimes DL models get stuck in local minima when these hyper-parameters are not optimized. Choras and Pawlicki \cite{article3} investigated the influence of the  hyper-parameters on the intrusion detection rate. There work was based on feed forward deep neural network (FFDNN) architecture which is one of the most popular DL model use in the development of IDS \cite{article306,article303,article301}. To find optimal hyper parameters, Zhang et al \cite{article305} improved a deep neural network for network intrusion detection using a genetic algorithm (GA). In their work, they used GA, to find the optimal number of hidden layers and the number of neurons in each layer to achieve a high detection rate. In the same vein, Benmessahel et al. \cite{article322} improved an artificial neural network (ANN) based model for intrusion detection using an evolutionary algorithm based optimizer. This results in avoiding local minima during ANN training and achieving high accuracy. Similarly, Wu et al \cite{article318} have proposed convolutional neural network (CNN) based IDS whose weights are adjusted based on the proportion of samples per class. This avoids bias towards the class with more samples in the training dataset and achieve better detection rate for minority intrusion class.
Kanna and Santh \cite{article5}  proposed a unified end-to-end learning model based on CNN and LSTM architecture. CNN learns spatial relationships while LSTM learns temporal features in data for feature extraction and classification. They also used Lion Swarm Optimization (LSO) to tune the CNN's hyperparameters for optimal spatial feature learning configuration.  Zhang et al.\cite{article4} considered temporal, spatial, and content associations as well as correlations between features to create multiple combinations of features that are used to train a stacking ensemble learning model (essentially a combination of multiple ML models) to detect intrusion attempts. Poornachandran et al. \cite{article302} proposed network intrusion detection system that models network traffic as time series and uses long-term short-term memory or identity recurrent neural  network for learning and identifying intrusion attacks from traffic packets modeled as predefined time segment. 

\section{Materials and methodology}
\begin{figure}[ht]
   \centering
    \includegraphics[scale=0.48]{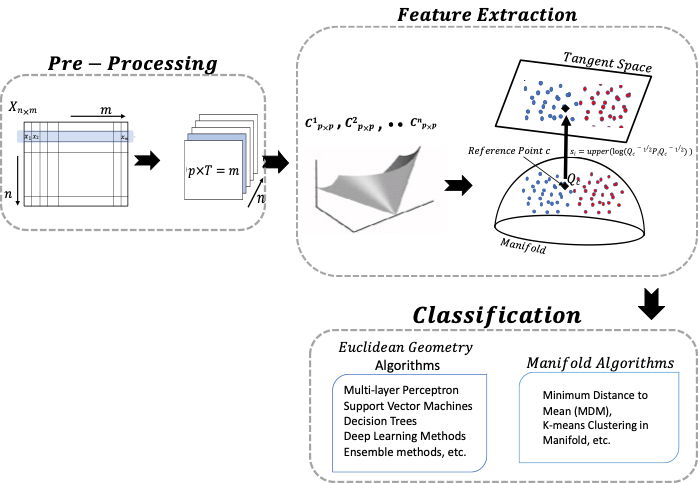}    
    \caption{A block diagram of our proposed approach}
    \label{fig:frameworks}
\end{figure}

Our proposed methodology for intrusion detection in the network traffic is presented in this section. Our proposed approach majorly consists of data preprocessing, feature extractions based on spatio-temporal properties of raw data and proposed approach for classification. Our methodology is illustrated in Figure ~\ref{fig:frameworks}.
\subsection{Dataset Description}
At present , only a few benchmark datasets can evaluate the effectiveness of the IDS model. To evaluate our proposed approach, we considered a well-known public datasets namely, NSL-KDD \cite{nslkdd} and UNSW-NB15 \cite{unswnb}. 
\subsubsection{Dataset I Description}
This benchmark dataset is freely available from the Canadian Institute of Cybersecurity. NSL-KDD dataset was obtained by removing redundant records from the KDDCUP99 dataset so that machine learning based models can produce unbiased results. In addition to normal traffic, this dataset consists of traffic from four attacks, namely DoS, U2R, R2L, and PROBE traffic.
\\
\begin{table}[ht]
\centering
\caption{NSL-KDD dataset}
\label{table:nslkdd2}
\scalebox{0.90}{
\begin{tabular}{@{}lcc@{}}
\toprule
$Class$  & NSL-KDD$_{Train+}$ & NSL-KDD$_{Test+}$ \\ \midrule
$Normal$  & 67343     & 9711     \\
$Attack$  & 57738     & 12833     \\
$Total$  & 125081     & 22544     \\ \bottomrule

\end{tabular}}
\end{table}
Table ~\ref{table:nslkdd2} shows count of samples from training and testing set from NSL-KDD dataset. As shown in the Table ~\ref{table:nslkdd2} is highly imbalanced with fewer instances from U2R and R2L attack classes. Furthermore, test dataset contains unknown attacks samples that do not appear in the training dataset as shown in Table ~\ref{table:nslkdd2}. 
Each traffic record in the NSL-KDD dataset is a vector of 41 continuous and nominal values. These 41 values can further subdivided into four categories. The first category is the intrinsic type, which essentially refers to the inherent characteristics of an individual connection. The second category contains indicators that relate to the content of the network connection. The third category receives a set of values based on the study of the content of the connections in the time segment of 2 seconds. Finally, the fourth category  is based on the destination host. 
\subsubsection{Dataset II Description}
UNSW-NB15 dataset is benchmark dataset that contains nine families of intrusion attacks, namely, Shellcode, 
Fuzzers, Generic, DoS, Backdoors, Analysis, Generic, Worms and Reconnaissance \cite{unswnb}. This dataset is freely provided by the Cyber Range Lab of the Australian Centre for Cyber Security (ACCS). We used already configured training and testing data set from ACCS as shown in Table \ref{table:UNSW}. The number of samples in training dataset is 175,341 and the test set has 82,332 from normal and nine type of attacks. The dataset has a total of 42 features and these features can be subdivided into categories. Similar to NSL-KDD dataset, first part is the  content features; second category has some features which refers to basic and general operation of the internet; third part is connection features and lastly fourth category is time based features.

\begin{table}[ht]
\centering
\caption{UNSW-NB15 dataset}
\label{table:UNSW}
\scalebox{0.90}{
\begin{tabular}{@{}lcc@{}}
\toprule
$Class$  & UNSW-NB15$_{Train}$ & UNSW-NB15$_{Test}$ \\ \midrule
$Normal$  & 56000     & 37000     \\
$Attacks$    & 119341    & 45332     \\ 
$Total$    & 175341    & 82332    \\ \bottomrule
\end{tabular}}
\end{table}
\subsection{Data Pre-processing}
In NSL-KDD dataset, there are some features that are symbolic and continuous. These features need to be converted into single numeric type for the feature extraction. Secondly, features are not uniformly distributed thus need to be scaled for better result with machine learning models.
\subsubsection{Data Standardization}
Data Standardization deals with the numeralization of categorical features. Most common method is encoding symbolic values with numeric values . For example, if feature contain three unique symbolic values like NSL-KDD $protocol$ $type$ contain $tcp$, $udp$ and $icmp$ then these attributes can be map with 1, 2 and 3 respectively.
\subsubsection{Data Normalization}
In general, attributes in traffic data are highly variable and not uniformly distributed. To achieve better results with the machine learning model, the attribute values are usually scaled to a uniform distribution in the interval $[0-1]$. For this purpose, the min-max normalization method is used, as shown in Equation ~\ref{eq:minmax}.

\begin{equation}\label{eq:minmax}
	x_{i}^{'} = \frac{x_{i} - min (x_i)}{max(x_i) - min(x_i)}
\end{equation}

Where $max(x_i)$ and $min(x_i)$ represent maximum and minimum value of feature vector $x_i$; whereas $x_{i}^{'}$ is a normalized feature value between [0-1].
\subsection{Feature Extraction}
\begin{figure}[ht]
    \centering
    \includegraphics[scale=0.60]{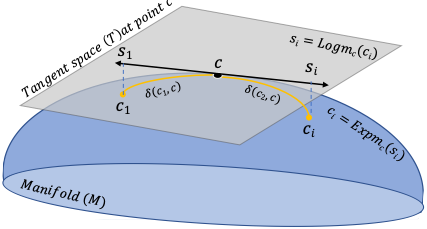}    
    \caption{An illustration of Riemannian manifold ($\mathcal{M}$)}
    \label{fig:framework}
\end{figure}

The traffic data $X \in \mathbb{R}^{n \times m}$  is a matrix of size  $n$ $\times$ $m$  where $n$ is the number of samples and  each sample $x_{i}$ = $\{x^1,x^2 ...., x^m\}$ is described by $m$ dimensional feature vector as illustrated in Figure \ref{fig:framework}. This $x_{i}$ $\in$ $\mathbb{R}^m$ sample can be reshape into $\mathbb{R}^{p\times t}$ to obtain
covariance matrix $C_{i}$  using equation \ref{eq:scm}:
\begin{equation} C_i=\frac{1}{t-1}{x_{i}x_{i}^T \in \mathbb{R}^{p \times p} }\label{eq:scm} \end{equation}
where $T (.)$ represent transpose operator of the matrix and $i$ is an index of the sample in data set ranging from $1$ to $N$. Covariance matrices encode spatio-temporal information and lie in the  symmetric positive definite (SPD) space. A set of all $p$ $\times$ $p$ square matrix $C$ ,  if $C= C^T$ and $\{u^TCu >0$, $\forall u \in \mathbb{R}^p, u\neq0\}$ then lies in the symmetric positive definite matrices space $SPD(p)$. Equivalently, SPD matrices are invertible i.e. $\forall C \in SPD(p), C^{-1} \in SPD(p)$. To make covariance matrix  invertible and well-conditioned (inverting does not amplify estimation error), we have used Ledoit-wolf regularization \cite{LEDOIT2004365} to estimate covariance matrix of sample as defined in equation \ref{eq:scm1}:
\begin{equation} \bar{C_i}= (1 - \alpha)C_i+ \alpha \frac{tr(C_i)}{p} Ip \label{eq:scm1} \end{equation}
where $tr(.)$ is trace operator, $\alpha$ shrinkage hyper-parameter and I is the identity matrix. SPD matrices forms a Riemannian hyper-cone $\mathcal{M}$ of dimension $p(p+1)/2$ \cite{1Alexandre}.\\
A Riemannian Manifold $\mathcal{M}$ is a differentiable topological space \cite{singh2019small}. The derivative of each point $c$ in Riemannian Manifold lies in tangent Euclidean vector space $\mathcal{T}$ as illustrated in Figure \ref{fig:framework}. The logarithmic mapping ($logm_{C}(C_i)$) operation is used to project any point (matrix) $C_i$ to tangent space $\mathcal{T}$. The equation \ref{eq:rie1} define logarithmic mapping operation:
\begin{equation}\small    S_i =logm_{C}(C_i)= C^{1/2}\log{({C^{-1/2}C_iC^{-1/2}})C^{1/2}} \label{eq:rie1}\end{equation}
where $C_i$ is a reference point to map tangent plane and $log(.)$ is logarithm of SPD matrix. Similarly, a tangent vector $S_i$ from tangent space $\mathcal{T}$ is projected back to manifold $\mathcal{M}$ using exponential mapping $expm_{C}(S_i)$. It is given by
\begin{equation}\small    C_i =expm_C(S_i)= C^{1/2} \exp{({C^{-1/2}S_iC^{1/2}})}C^{1/2} \label{eq:rie11}\end{equation}
where $exp(.)$ is exponential operation of SPD matrix. These logarithm and exponential mapping are illustrated in figure \ref{fig:framework}.
The Riemannian manifold $\mathcal{M}$ is equipped with  distance (geodesic) metric that is unique and shortest path connecting two points $C_1$ and $C_2$, it is given by
\begin{equation} \footnotesize  \delta_G(C_1,C_2)=\| \log(C_1^{-1/2}C_2C_1^{-1/2})\|_F=(\sum_{i=1}^n\log^2\lambda_i)^{1/2} \label{eq:rie2}\end{equation}
where $\lambda_i$'s are the positive eigenvalues of $C_1^{-1/2}C_2C_1^{-1/2}$ and $\|.\|_F$ denotes the Frobenius norm of the matrix. 
Riemannian (geodesic) distance between two points depends upon the best approximation of reference point $C$ on which tangent space is mapped. Usually, mean $(Q_{c})$  is used as reference point. It is a unique point in the manifold $\mathcal{M}$ that minimized the sum of squared distances between projected SPD matrices and is given by equation \ref{eq:rie3}.
\begin{equation} Q_{c}(C_1,...,C_n)=\underset{{C \in SPD(p)}}{\arg \min} {\sum_{i=1}^{N}{\delta^2(C,C_i)}}  \label{eq:rie3} \end{equation}
After $Q_{c}$ mean is computed, each SPD matrix $C_i$ can be projected on the tangent space and vectorized as:
\begin{equation} S_i= upper(\log{(Q_{c}^{-1/2}C_i Q_{c}^{-1/2})} ) \label{eq:tsm1} \end{equation}
Where is $S_i$ is tangent space vector of $p(p+1)/2$ dimension and $upper (.)$ is an operator that keeps upper triangular part of a symmetric matrix and vectorizes it \cite{Islam_2018}. This is illustrated in figure \ref{fig:framework}. The tangent space mapped (TSM) features of traffic data can be denoted as V $\in$ $\mathbb{R}^{n \times (p(p+1)/2)}$:
\begin{equation} V = [s_1, s_2, ...., s_n]^T   \label{eq:tsm2} \end{equation}
\begin{figure}[ht]
	\centering
	\includegraphics[width=0.85\linewidth]{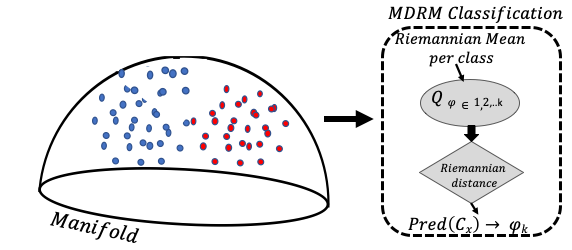}
	\caption{Illustration of MDRM on Riemannian manifold}
	\label{fig:mdrm}
\end{figure}
\subsection{ Riemannian Manifold based Classification}
Manifold based classification methods works directly on SPDs for classification. One such method is Minimum distance to Riemannian mean (MDRM) that utilizes the Riemannian mean of each class (normal traffic or abnormal traffic) and its Riemannian distance to predict class label for new sample as illustrated in figure \ref{fig:mdrm}. In order to do so, firstly using labelled training samples  Riemannian mean is obtained for each class, then  new traffic sample's covariance matrix  distance (Riemannian) with respect to each class is obtained. The shortest Riemannian distance determine new sample is nearest to particular class mean. Thus, that class becomes predicted label for new sample as depecited in equation \ref{eq:mdrm1}.
\begin{equation}pred(C_x)=\underset{\varphi=1,2..K}{\arg \min} R_d(C_x,Q_\varphi) \label{eq:mdrm1} \end{equation}

where $Q_\varphi$  denotes the Riemannian mean of Class $\varphi$, $C_x$ represent the covariance matrix of the new sample and $pred(C_x)$ is the prediction of its class label.   



\subsection{ Euclidean space based Classification}
Most of the machine learning and deep learning classification algorithm works in Euclidean space. Thus, to exploit geometrical properties, tangent space mapped features can be obtained from covariance matrices. In our work, we employed two deep learning and three shallow learning Euclidean space classification methods namely  Multi-layer Perceptron (MLP) , autoencoder (AE), support vector machines (SVM),  linear discriminant analysis (LDA) and ensemble learning  algorithm. All these methods use tangent space mapped (TSM) features (V $\in$ $\mathbb{R}^{n \times (p(p+1)/2)}$) for classification.
\begin{figure}[ht]
	\centering
	\includegraphics[width=0.85\linewidth]{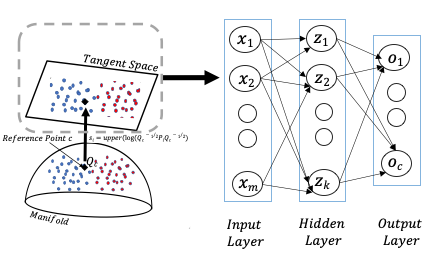}
	\caption{Illustration of Multi-layer perceptron with TSM features}
	\label{fig:mlp}
\end{figure}
The Multilayer perceptron (MLP)  \cite{Goodfellow-et-al-2016} is a feed forward neural network having multiple hidden layers, one input layer and output layer and can be represented as a directed graph without forming any cycle as shown in Figure ~\ref{fig:mlp}. The tangent space feature vector is passed through input layer (size equal to TSM feature vector) whose information is extracted through multiple hidden layers containing fully connected neurons.  
Finally, output layer (size equal to number of classes) based on non-linear activation function (mostly sigmoid or softmax) predict malicious network traffic.
Similarly, an autoencoder (AE) like MLP is also a feed forward neural network  composes of an input layer, an output layer, and one or more hidden layers. Unlike MLP, AE have input layer and output layer of same size. In our study, it be equal to dimension of tangent space feature vector.
Once the AE model is trained with normal traffic's then any traffic above threshold  (usually two standard deviation away from reconstruction error on normal samples) is labeled as malicious. In this study , 5-layer  architecture with the number of neurons in each layer equal to [TSM-32-5-32-TSM] is used to build AE model. Here, TSM  indicates  size of input layer equal to dimension of TSM feature vector.\\
In contrast to above two methods, Linear Discriminant Analysis (LDA)  is a shallow learning supervised method mostly used for classification and dimensionality reduction \cite{duda2006pattern}. LDA finds a projection vector that maximises the distance between classes  and minimizes the variance within a class. 
Support vector machine (SVM) is also statistical method that finds the best hyper-plane or multiple hyper-planes that provides the maximum separation between classes \cite{steinwart2008support}. 
In our study, a SVM classifier with linear kernal is used for classification. A detailed mathematical formulation of the SVM is reported in \cite{steinwart2008support} . Lastly, 
ensemble learning (EL) is a  meta learning machine learning approach where to better performance is achieved by combining multiple machine learning models \cite{sensethepen}.
In this study, we have used extra tree (ETC) ensemble learning model which is based on decision trees (DT) as a base ML model. 
\section{Experiment setup and evaluation metrics}
This section describes experiment setup, performance metrics and the dataset used to evaluate the proposed approach. 
\subsection{Experimental Setup}
This study was carried out using 2.3 GHz 8-core Intel i9 processor with 16 GB memory on MacOS Big Sur 11.4 operating system. The proposed approach is developed using Python programming language with several statistical and visualization packages such as Sckit-learn, Numpy, Pandas, Tensorflow and Matplotlib. Table~\ref{table:Mat} summarizes the system configuration.
\begin{table}[ht]
	\centering
	\footnotesize
	\caption{Implementation environment specification}
	\label{table:Mat}
	\begin{tabular}{p{2.6cm} | p{3.8cm}}
		\hline
		\textbf{Unit}   & \textbf{Description}\\ \hline
		Processor   & 2.3 GHz 8-core Inter Core i9 \\ \hline
		RAM  &  16 GB      \\ \hline
		Operating System  &  MacOS Big Sur 11.4  \\ \hline	
		Packages   &  Tensorflow, Sckit-Learn, Numpy, Pandas, Pyriemannian and Matplotlib    \\ \hline
	\end{tabular}
\end{table}
\subsection{Evaluation Metrics}
The proposed method is compared and evaluated using Accuracy, Precision, Recall, F1-score and Area under the receiver operating characteristics (ROC) curve. In this work, we have used the macro and micro average of Recall, Precision and F1-score for multi-class classification. All the above metrics can be obtained using the confusion matrix (CM). The Table ~\ref{tab:conmat} illustrates CM for binary classes, but can be extended to multiple classes.\\
\begin{table}[h]
\centering
\caption{Illustration of confusion matrix}
\label{tab:conmat}
\scalebox{0.85}{
\begin{tabular}{@{}cccc@{}}
\toprule
\multicolumn{1}{l}{}    &         & \multicolumn{2}{c}{Predicted}        \\\cmidrule(l){2-4}  
\multicolumn{1}{l}{}    &         & Class$_{pos}$ & Class$_{neg}$ \\ \midrule
\multirow{2}{*}{Actual} & Class$_{pos}$ & True Positive $(TP)$     & False Positive $(FP)$      \\
                        & Class$_{neg}$ & False Negative $(FN)$     & True Negative $(TN)$      \\
\bottomrule
\end{tabular}}
\end{table}
In Table~\ref{tab:conmat}, True positive (TP) means amount of class$_{pos}$ data predicted actual belong to class$_{pos}$, True negative (TN) is amount of class$_{neg}$ data predicted is actually class$_{neg}$, False positive (FP) indicates data predicted class$_{pos}$ is actual belong to class$_{neg}$ and False negative (TN) is data predicted as class$_{neg}$ but actually belong to class$_{pos}$. Based on the aforementioned terms, the evaluation metrics are calculated as follows. \\
Accuracy (ACC) measures the total number of data samples are correctly classified as shown in equation ~\ref{eq:ACC}. For balanced test dataset, higher accuracy is indicate model is well learned, but for unbalanced test dataset scenarios relying on accuracy can give wrong illusion about model's performance.
\begin{equation}\label{eq:ACC}
	ACC = \frac{TP+TN}{TP + TN + FP + FN}
\end{equation}
Recall (also known as true positive rate) estimates the ratio of the correctly predicted samples of the class to the overall number of instances of the same class. It can be computed using equation ~\ref{eq:TPR}. Higher Recall $\in [0,1]$ value indicate good performance of machine learning model.
\begin{equation}\label{eq:TPR}
	Recall = \frac{TP}{TP + FN}
\end{equation}
Precision measures the quality of the correct predictions. mathematically, it is the ratio of correctly predicted samples to the number of all the predicted samples for that particular class as shown in equation equation~\ref{eq:PPV}. Precision is usually paired with Recall to evaluate performance of model. Sometime pair can appear contradictory thus comprehensive measure F1-score is considered for unbalanced test data-sets.
\begin{equation}\label{eq:PPV}
	Precision = \frac{TP}{TP + FP}
\end{equation}
F1-Score computes the trade-off between precision and recall. Mathematically, it is harmonic mean of precision and recall as shown in equation ~\ref{eq:F-measure}.
\begin{equation}\label{eq:F-measure}
	F = 2\times\left(\frac{Precision\times Recall}{Precision + Recall}\right)
\end{equation}
Area under the curve (AUC) computes the area under the receiver operating characteristics (ROC) curve which is plotted based on the trade-off between the true positive rate on the y-axis and false positive rate on the x-axis across different thresholds. Mathematically, AUC is computed as shown in equation ~\ref{eq:auc}.
\begin{equation}\label{eq:auc}
AUC_{ROC}=\int_{0}^{1} \frac{TP}{TP+FN}d\frac{FP}{TN+FP}
\end{equation}

\begin{figure}[ht]
	\centering
	\includegraphics[width=0.75\linewidth]{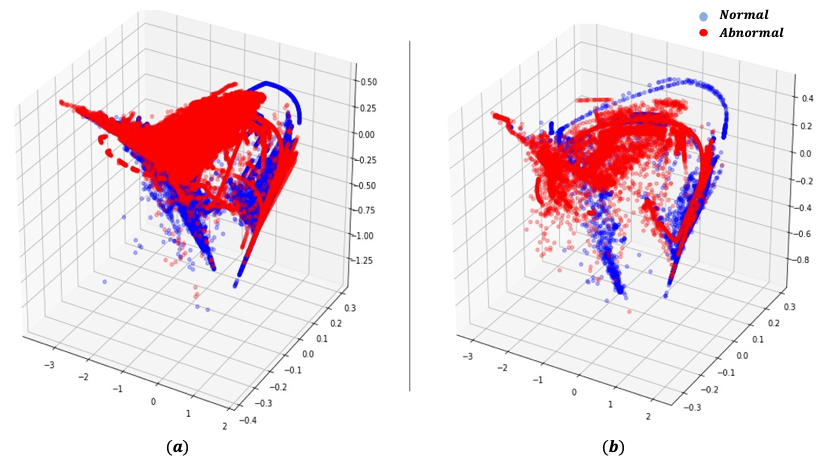}
	\caption{Tangent space visualization: (a) shows original kddTrain+ dataset, (b) shows Kddtest+ dataset}
	\label{fig:TSmm}
\end{figure}
\begin{figure}[ht]
	\centering
	\includegraphics[width=0.78\linewidth]{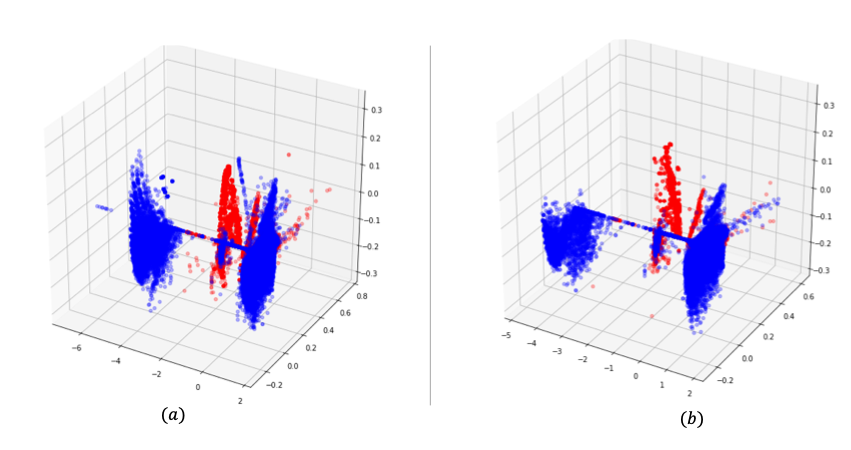}
	\caption{Tangent space visualization of UNSW-NB15: (a) shows original training dataset, (b) shows testing dataset respectively}
	\label{fig:TSmmu}
\end{figure}

\section{Results and Discussion}
\begin{figure}[ht]
	\centering
	\includegraphics[width=0.85\linewidth]{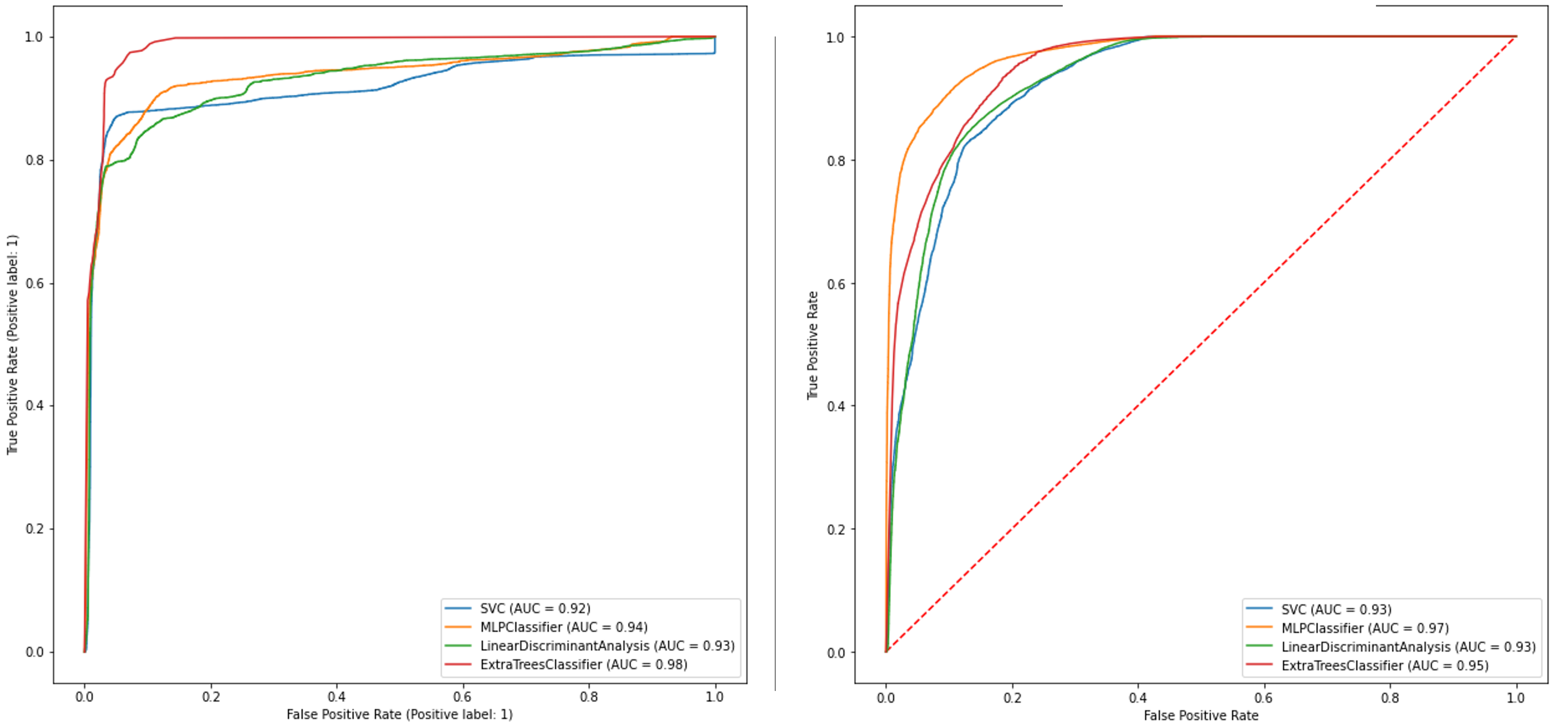}
	\caption{Receiver operating characteristic graph showing area under curve for different euclidean classifiers using tangent space features}
	\label{fig:auc1}
\end{figure}

We have made a number of different observations to understand the performance implications both during the training and testing phases.
The visualization of the normal and  abnormal  samples  across  KDDTrain+  and  KDDTest+ in  the  tangent  space is  shown  in  Figure \ref{fig:TSmm}.  The  tangent space here is obtained through $2 \times 2$ matrices, therefore lies in the three dimensional space. These $2 \times 2$ matrices are obtained by reshaping the raw data into $2 \times 20$ data matrices. We can clearly see two clusters in three dimensional space in  Figure \ref{fig:TSmm}. Similar is the case with UNSW-NB15 dataset, there are two clusters for normal and malicious traffic's tangent space mapped feature space as shown in Figure \ref{fig:TSmmu}. The tangent space mapping  feature space in Figure \ref{fig:TSmmu} is obtained using $3 \times 3$ matrices (lies in six dimensional space) by reshaping the raw data (we only used numeric features) into $3 \times 13$ data matrix.\\
\begin{figure}[ht]
	\centering
	\includegraphics[width=0.85\linewidth]{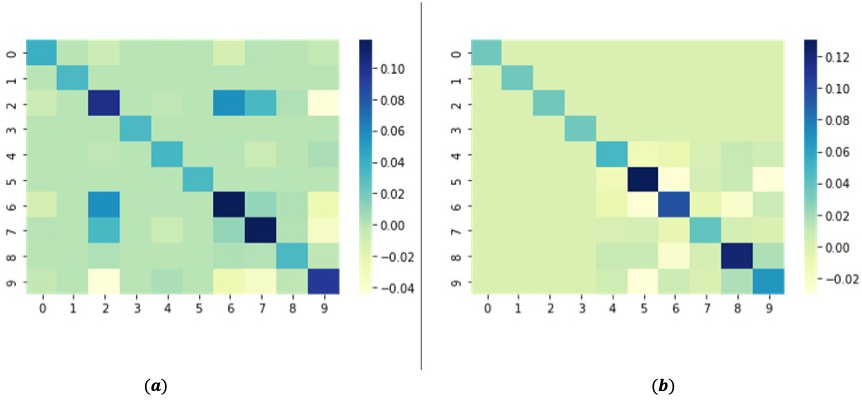}
	\caption{Mean covaraiance matrices visualisation: (a) Mean Covariance matrix of normal traffic samples, (b) Mean Covariance matrix of abnormal traffic samples}
	\label{fig:means}
\end{figure}
\begin{figure}[ht]
	\centering
	\includegraphics[width=0.85\linewidth]{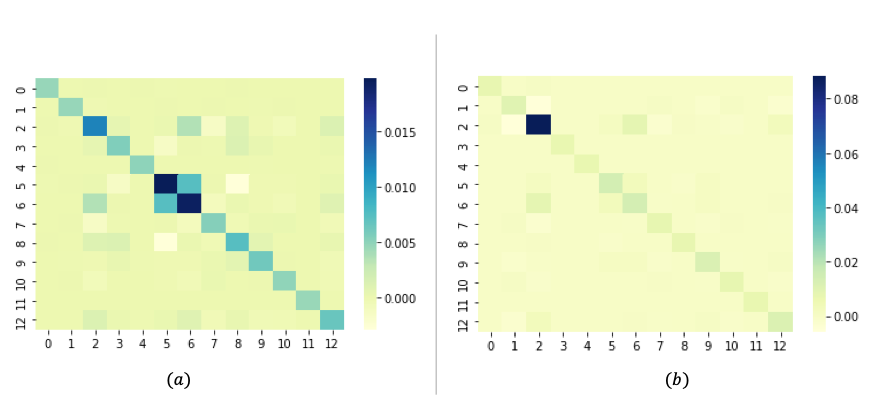}
	\caption{Mean covaraiance matrices visualisation of UNSW-NB15 dataset : (a) Mean Covariance matrix of normal traffic samples, (b) Mean Covariance matrix of abnormal traffic samples}
	\label{fig:meansu}
\end{figure}
We have used $10 \times 4$ matrix matching original four categories of NSL-KDD dataset and $13 \times 3$ reshaped matrix in UNSW-NB15 dataset to obtain covariance matrices  of size $10 \times 10$ and $13 \times 13$ respectively for classification. Figure \ref{fig:means} shows heat-map visualization of Riemannian means obtained from normal and malicious traffic samples covariance matrices in NSL-KDD dataset. We can clearly see different mutual relationships between variables for normal and abnormal samples. These mean covariance matrices are uses for the binary classification in manifold as new unseen sample is labelled based on its distance from covariance matrix shown in figure \ref{fig:means}. Similarly, figure \ref{fig:meansu} shows mutual relationship in UNSW-NB15 dataset and their we can clearly same trend here in the UNSW-NB15 as it was with NSL-KDD. There are a lot of dependencies between different variable in normal traffic compare to malicious traffic data.
\begin{table}[ht]
\centering
\footnotesize
\caption{ Performance comparison of different classification approaches on two datasets}
	\label{table:compare3}
\begin{tabular}{@{}llllllll@{}}
\toprule
\multirow{2}{*}{Datasets}  & \multirow{2}{*}{\begin{tabular}[c]{@{}l@{}}Performance\\ Metrices\end{tabular}} & \multicolumn{6}{c}{} \\ \cmidrule(l){3-8} 
                           &                                                                                 & MDRM   & SVM   & LDA   & EL   & AE     & MLP  \\ \cmidrule(l){2-8} 
\multirow{4}{*}{NSL-KDD}   & Accuracy                                                                        & 86     & 84    & 86    & 88   & 92     & 87   \\
                           & Precision                                                                       & 88     & 87    & 88    & 89   & 92.33  & 89   \\
                           & Recall                                                                          & 86     & 84    & 86    & 87   & 91.12  & 87   \\
                           & F-score                                                                         & 86     & 84    & 86    & 87   & 92.26  & 87   \\ \cmidrule(l){2-8} 
\multirow{4}{*}{UNSW-NB15} & Accuracy                                                                        & 74     & 82    & 81    & 88   & 88.67  & 89   \\
                           & Precision                                                                       & 78     & 87    & 87    & 89   & 87     & 89   \\
                           & Recall                                                                          & 73     & 82    & 81    & 86   & 86     & 87   \\
                           & F-score                                                                         & 71     & 81    & 81    & 86   & 86     & 87   \\ \bottomrule
\end{tabular}
\end{table}
Table \ref{table:compare3} shows weighted average performance metrices of different classification approaches with Riemannian manifold based feature extraction in NSL-KDD and UNSW-NB15 datasets respectively. Figure \ref{fig:auc1} is receiver operating characteristic graph showing area under curve for different Euclidean classification method for both datasets.

\begin{table}[ht]
\centering
\footnotesize
\caption{ Performance comparison with other state of the art approaches on KDDTest+}
	\label{table:compare}
\setlength{\tabcolsep}{1.6mm}
 \begin{tabular}{lrrrr} 
\hline
Method & Accuracy  & Precision & Recall & F1-score \\
\hline
ANN \cite{7058223} & 81.2 & \textbf{96.59} & 69.35 & 80.73\\
Sparse-AE + SVM \cite{8463474}  & 84.96 & \textbf{96.23} & 76.57 & 85.28 \\
AE \cite{IERACITANO202051}  & 84.21 & 87 & 80.37 & 81.98 \\
LSTM\cite{IERACITANO202051}  & 82.04 & 85.13 & 77.70& 79.24\\ 
MLP\cite{IERACITANO202051}  & 81.65 & 85.03 & 77.13 & 78.67 \\
L-SVM\cite{IERACITANO202051}  & 80.8 &84.32& 76.06 & 77.54 \\
Q-SVM\cite{IERACITANO202051}  & 83.15 & 86.09& 79.86 & 81.39\\
LDA\cite{IERACITANO202051} & 79.27 & 84.09 & 73.81 & 75.16 \\
QDA\cite{IERACITANO202051}  & 76.84 & 78.71 & 77.23 & 77.78 \\
\textbf{Our method} & \textbf{92.33} & 92.4 & \textbf{91.12} & \textbf{92.26}\\ 
\hline
\end {tabular}
\end{table}

\begin{table}[ht]
\centering
\footnotesize
\caption{Performance comparison with other approaches on UNSW-NB15}
	\label{table:comparessnb}
\setlength{\tabcolsep}{1.6mm}
 \begin{tabular}{lrrrr} 
\hline
Method & Accuracy  & Recall   & Precision& F1-score \\
\hline
CatBoost \cite{article4} & 87.40& \textbf{73.92} & 97.43 & 84.06\\
LightGBM\cite{article4}  & 87.61 & \textbf{74.14} & 97.76 & 84.33 \\
AdaBoost \cite{article4}  & 86.41 & 72.83 & 95.96 & 82.81\\
Extra Trees\cite{article4}  & 86.73 & 72.60 & 97.14 & 83.10 \\ 
Voting\cite{article4}  & 86.15 & 74.71 & 93.12 & 82.90 \\
MFFSEM \cite{article4}  & 88.85 & 80.44 & 93.88 & 86.64 \\
\textbf{Our method} & \textbf{89} & \textbf{87} & 89 & \textbf{87}\\ 
\hline
\end {tabular}
\end{table}
We have also compared the performance of our best performing model based on Manifold feature extraction with other state of the art models in the literature by using the four metrics namely accuracy, precision, recall, and F1-score. The table \ref{table:compare} shows that our best performing proposed (TSM features based AE model for NSL-KDD) obtained high recall and F-score respectively compared to other state of the art methods. In anomaly detection, more than accuracy, high recall and F-score respectively indicates quality of the model.
Similarly, We compared our best performance model (MLP for UNSW-NB15) based on tangent space features with other state of the art methods in the literature for UNSW-NB15 dataset. Table \ref{table:comparessnb} shows our model obtained high recall and F-score respectively compared to other state of the art methods. All our classification models used in this work are with default parameters provided by sklearn library. If we fine-tune our models parameters with some optimization technique, we can further increase the performance of our intrusion detection models which is mostly the case with other state of the art methods in the literature \cite{article4}.

\section{Conclusion}
In this paper, a novel feature extraction method is proposed that captures spatial and temporal characteristics of data. our IDS model uses covariance matrices and their geometrical properties to detect  intrusion attacks. Moreover, the proposed approach exploits tangent space mapping that inherits spatio-temporal and geometrical information of covariance matrices to use with any state of the art Euclidean space algorithms for intrusion detection. The performance of the proposed approach was evaluated using the NSL-KDD and UNSW-NB15 datasets respectively. The experimental results show that the proposed approach not only has good detection performance , but also outperform the existing IDS models. The proposed approach can be further improved by using different feature selection techniques as well as augmenting minority attacks in the dataset. In future research, we will try to use conditional Generative adversarial networks  along with tangent space features to detect minority attacks. Another direction need to be investigated is unsupervised intrusion attack detection in the manifold. As in the real world, it is very impractical task to label normal and intrusion data packets.
\section*{Acknowledgment}
This research is supported by the Cyber Security Research Programme—Artificial Intelligence for Automating Response to Threats from the Ministry of Business, Innovation, and Employment (MBIE) of New Zealand as a part of the Catalyst Strategy Funds under the
grant number MAUX1912.

\bibliography{Bibliography}

\end{document}